%% file: ms.tex
\pdfoutput=1

\documentclass[11pt]{article}

\usepackage[final]{naacl2021}

\usepackage{times}
\usepackage{latexsym}

\usepackage[T1]{fontenc}

\usepackage[utf8]{inputenc}

\usepackage{microtype}

\usepackage{booktabs}
\usepackage{subcaption}

\usepackage{graphicx}
\usepackage{mkolar_definitions}
\usepackage{multirow}

\usepackage{algorithm}
\usepackage[noend]{algorithmic}

\usepackage{fancyvrb} 
\usepackage{xspace}

\newcount\Comments  
\definecolor{darkgreen}{rgb}{0,0.5,0}
\definecolor{darkred}{rgb}{0.7,0,0}
\definecolor{teal}{rgb}{0.3,0.8,0.8}
\definecolor{blue}{rgb}{0,0,1}
\definecolor{purple}{rgb}{0.5,0,1}
\newcommand{\kibitz}[2]{\ifnum\Comments=1\textcolor{#1}{#2}\fi}

\newcommand*{\affaddr}[1]{#1} 
\newcommand*{\affmark}[1][*]{\textsuperscript{#1}}
\newcommand*{\email}[1]{\texttt{#1}}

\title{Clustering-based Inference for Biomedical Entity Linking}

\author{%
Rico Angell\affmark[1], Nicholas Monath\affmark[1], Sunil Mohan\affmark[2], Nishant Yadav\affmark[1], and Andrew McCallum\affmark[1]\\
\affaddr{\affmark[1] College of Information and Computer Sciences \\ University of Massachusetts Amherst}\\
\email{\{rangell,nmonath,nishantyadav,mccallum\}@cs.umass.edu}\\
\affaddr{\affmark[2]Chan Zuckerberg Initiative}\\
\email{smohan@chanzuckerberg.com}\\
}

\begin{document}
\maketitle
\input{notation}

\input{01_abs}
\input{02_intro}
\input{03_problem}

\input{04_model}

\input{05_training}
\input{07_experiments}

\input{08_related_work}

\input{09_conclusion}
\input{11_ethics}
\input{10_ack}

\bibliography{anthology,references}
\bibliographystyle{acl_natbib}


\end{document}

%% file: notation.tex
\newcommand{\entity}{\ensuremath{e}}
\newcommand{\mention}{\ensuremath{m}}

\newcommand{\cC}{\mathcal{C}}
\newcommand{\cE}{\mathcal{E}}
\newcommand{\cL}{\mathcal{L}}
\newcommand{\cM}{\mathcal{M}}
\newcommand{\cD}{\mathcal{D}}

\newcommand{\R}{\mathbb{R}}
\newcommand{\gte}{\ensuremath{e^\star}}
\newcommand{\pe}{\ensuremath{\hat{e}}}

%% file: 01_abs.tex
\begin{abstract}
    Due to large number of entities in biomedical knowledge bases, only a small fraction of entities have corresponding labelled training data.
    This necessitates entity linking models
    which are able to link mentions of unseen entities
    using learned representations of entities.
    Previous approaches link each mention independently, ignoring the relationships within and across documents between the entity mentions. These relations can be very useful for linking mentions in biomedical text where linking decisions are often 
    difficult due mentions having a generic or
    a highly specialized form. 
    In this paper, we introduce a model in which 
    linking decisions can be made not merely by linking to 
    a knowledge base entity but also by grouping multiple mentions together via
    clustering and jointly making linking predictions. 
    In experiments on the largest publicly available biomedical dataset, we improve 
    the best independent prediction for
    entity linking by 3.0 points of accuracy, and
    our clustering-based inference model further improves entity linking by 2.3 points.
\end{abstract}

%% file: 02_intro.tex
\section{Introduction}
Ambiguity is inherent in the way entities are mentioned in natural language text. Grounding such ambiguous mentions to their corresponding entities, the task of \emph{entity linking},  is critical to many applications: automated knowledge base construction and completion \cite{riedel2013relation,surdeanu2012multi}, information retrieval \cite{Meij:2014:ELR:2556195.2556201}, smart assistants  \cite{balog2019personal}, question answering  \cite{dhingra2020differentiable}, text mining \cite{leaman2016taggerone,murty2018hierarchical}.

\begin{figure}[t]
    \centering
    \includegraphics[width=0.5\textwidth]{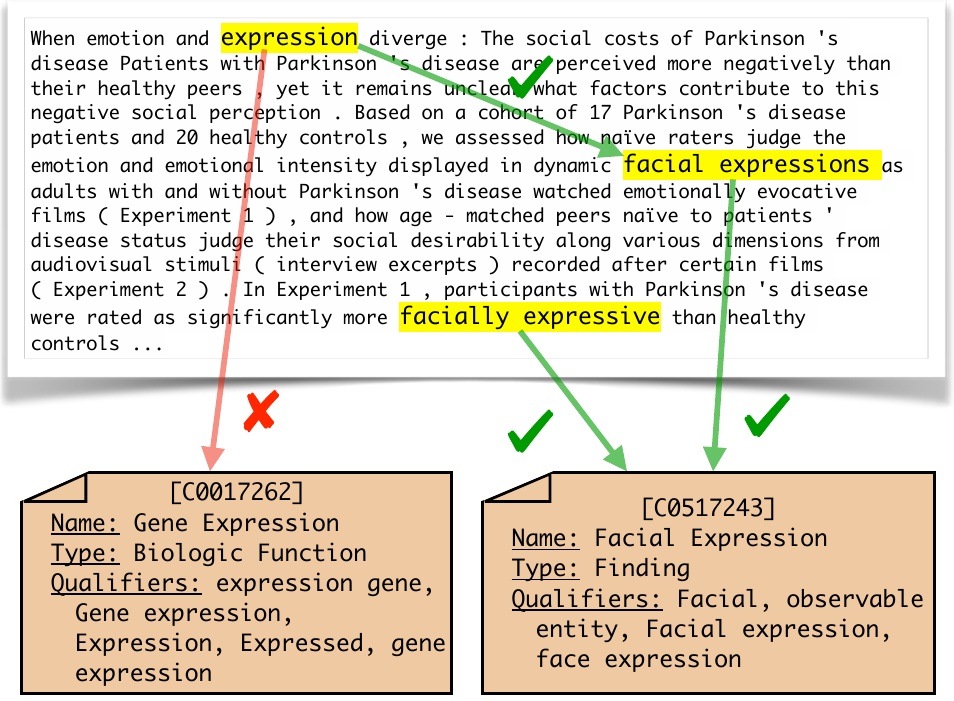}
    \caption{\textbf{Biomedical Entity Linking}.  All three highlighted mentions refer to the same entity. The mention \texttt{expression} is clearly related to the other two highlighted mentions which are much less ambiguous. If considered independently \texttt{expression} is more closely related to an incorrect entity.}
    \label{fig:motivating_example}
\end{figure}

Consider the excerpt of text from a biomedical research paper in Figure \ref{fig:motivating_example}, the three highlighted mentions (\emph{expression}, \emph{facial expressions}, and  \emph{facially expressive}) all link to the same entity, namely \texttt{C0517243 - Facial Expresson} in the leading biomedical KB, Unified Medical Language System (UMLS).

The mention \emph{expression} is highly ambiguous and easily confused with the more prevalent entity, \texttt{Gene expression}. This linking decision may become easier with sufficient training examples (or sufficiently rich structured information in the knowledge-base) . However, in biomedical \cite{mohan2019medmentions} and other specialized domains \cite{logeswaran2019zero}, it is often the case that the knowledge-base information is largely incomplete. Furthermore, the scarcity of training data leads to a setting in which most entities have not been observed at training.

State-of-the-art entity linking methods which are able to link entities unseen at training time make predictions for each mention independently \cite{logeswaran2019zero,wu2019zero}.
In this way, the methods may have difficulty linking mentions 
which, as in the example above, have little lexical similarity with the entities in the knowledge-base, as well as mentions for which the context is highly ambiguous. These mentions cannot directly use information from one mention (or its linking decision) to inform the prediction of another mention. On the other hand, entity linking methods that do jointly consider entity linking decisions  \cite{ganea2017deep,le2018improving} are designed for cases in which all of the entities in the knowledge-base to have example mentions or meta-data at training time \cite{logeswaran2019zero}.

In this paper, we propose an entity linking model in which entity mentions are either (1) linked directly to an entity in the knowledge-base or (2) join a cluster of other mentions and link as a cluster to an entity in the knowledge-base. Some mentions may be difficult to link directly to their referent ground truth entity, but may have very clear coreference relationships to other mentions. So long as one mention among the group of mentions clustered together links to the correct entity the entire cluster can be correctly classified. This provides for a joint, tranductive-like, inference procedure for linking.
We describe both the inference procedure as well as training objective for optimizing the model's inference procedure, based on recent work on supervised clustering \cite{yadav2019supervised}. 

It is important to note that our approach does not aim to do joint coreference and linking, but rather makes joint linking predictions by clustering together mentions that are difficult to link directly to the knowledge-base.
For instance, in Figure \ref{fig:motivating_example}, the mention \emph{expression} may be difficult to link to the ground truth \texttt{Facial expression} entity in the knowledge-base because the mention can refer to a large number of entities. However, the local syntactic and semantic information of the paragraph give strong signals that \emph{expression} is coreferent with \emph{facial expression}, which is easily linked to the correct entity. 

We perform experiments on two biomedical entity linking datsets:
MedMentions \cite{mohan2019medmentions}, the largest publicly available dataset as well as the benchmark BC5CDR \cite{li2016biocreative}. We find that our approach improves over our strongest baseline by 2.3 points of accuracy on MedMentions and 0.8 points of accuracy on BC5CDR over the baseline method \cite{logeswaran2019zero}. We further analyze the performance of our approach and observe that (1) our method better handles ambiguous mention surface forms (as in the example shown in Figure~\ref{fig:motivating_example}) and (2) our method can correctly link mentions even when the candidate generation step fails to provide the correct entity as a candidate.


%% file: 03_problem.tex
\section{Background}

Each document $D \in \cD$, has a set of mentions
$\cM^{(D)} = \{m_1^{(D)}, m_2^{(D)}, \ldots, m_N^{(D)}\}$. The task of entity linking 
is to classify each mention $m_i$ as referent to a single entity $e_i$ from a KB of entities. We use $\cE(m_i)$ to refer to the ground truth entity of mention $m_i$ and $\pe_i$ to refer to the predicted entity. 

\noindent \textbf{Knowledge-bases}. We assume that we are given 
a knowledge-base corresponding to a closed world of entities. These KBs are typically massive: English Wikipedia contains just over 6M  entities\footnote{number of content pages as of May 20, 2020, \url{https://en.wikipedia.org/wiki/Special:Statistics}} and the 2020 release of the  UMLS contains 4.28M entities\footnote{\url{https://www.nlm.nih.gov/research/umls/knowledge_sources/metathesaurus/release/notes.html}}. We describe in Sections~\ref{sec:dataset} \& \ref{subsec:bc5cdr} the details of the KBs used in each of the experiments.

\noindent \textbf{Candidate Generation}. Given the massive number of entities to which a mention may refer, previous work \cite[inter alia]{logeswaran2019zero} uses a candidate generation step to reduce the restrict the number of entities considered for a given mention, $m$, to a candidate set $\Gamma(m)$. The recall of this step is critical to the overall performance of entity linking models. 



%% file: 04_model.tex
\section{Model}
\begin{figure*}[t!]
    \centering
    \includegraphics[scale=0.25, trim=0 0 0 0, clip]{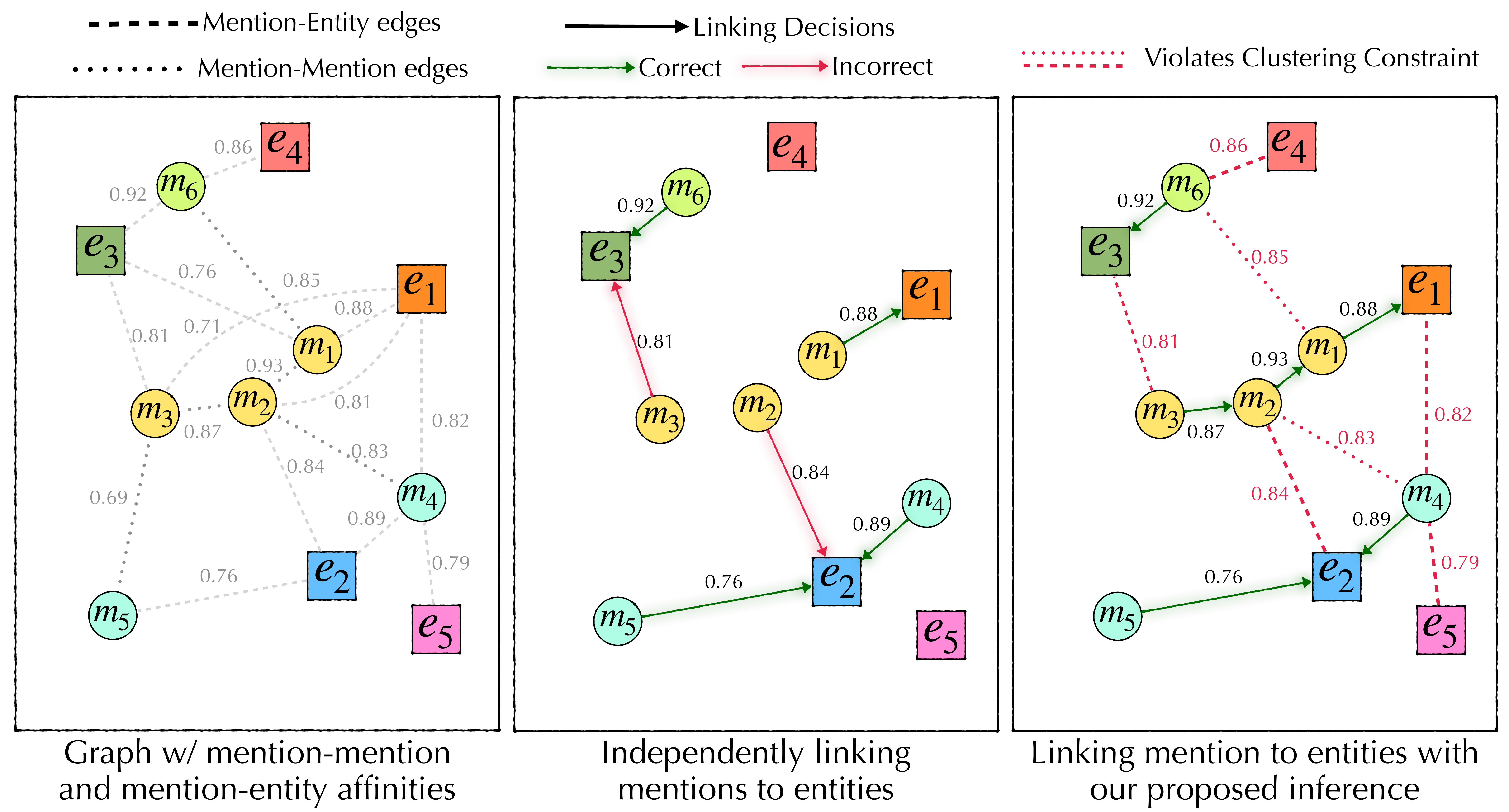}
    \caption{\textbf{Clustering-based Inference for Entity Linking}. Mentions are shown in circles and entities in squares. Color families indicate ground-truth cluster assignments. The left figure shows the graph $G$ that is the basis of the clustering task, the center figure show predictions under independent linking model, and  the right figure shows our proposed inference linking mentions to entities by running a constrained clustering inference procedure over $G$ that assigns at most one entity per cluster.}
    \label{fig:inference}
\end{figure*}

\label{sec:model}
In this section, we describe our clustering-based approach 
for jointly making entity linking predictions for a set of mentions. 
Our proposed inference method builds a graph where the nodes are the 
union of all of the mentions and entities and the edges have weights 
denoting the affinities between the endpoints. 
To make linking decisions, we cluster the nodes of the graph such that  each cluster contains exactly one entity, following which  
each mention is assigned to the entity in its cluster.

\subsection{Clustering-based Entity Linking}
Let $\varphi: \cM \times \cM \to \R$ and $\psi: \cM \times \cE \to \R$ be parameterized functions which compute mention-mention and mention-entity affinities, respectively.
The exact parameterizations of these functions are detailed in Section~\ref{sec:affinity_models}. 

Define the graph $G = (V, E)$ where $V = \cM \cup \cE$ and $E = \cM \times \cM \cup \{(m, e) : e \in \Gamma(m)\}$. The weight of each edge, $w(v_i, v_j)$ for $v_i, v_j \in V$, is determined by $\varphi$ or $\psi$ depending on the vertices of the edge: $w(m_i, m_j) = \varphi(m_i, m_j)$ and $w(m_i, e_l) = \psi(m_i, e_l)$. 
Linking decisions for each mention are determined by clustering the vertices of $G$ under the constraint that every entity must appear in exactly one cluster.

Given the graph $G$, we start with every node in their own individual cluster. We define affinity between a pair of clusters as the strongest cross-cluster edge between nodes in the two clusters. Iteratively, we greedily merge clusters by choosing a pair of clusters with largest affinity between them under the constraint that we cannot merge two clusters which \emph{both} contain an entity. When every cluster contains exactly one entity, this process can no longer merge any clusters, and thus terminates\footnote{This process is equivalent to single-linkage hierarchical agglomerative clustering with the constraint that two entities cannot be in the same cluster.}. Each mention is linked to the entity present in its cluster at the end of inference.
Algorithm~\ref{alg:clustering} describes this process of constructing the graph and clustering nodes to make linking decisions more formally.

Figure~\ref{fig:inference} shows the proposed inference in action on five entities and six mentions. Initially, every mention and entity start in a singleton cluster. 
In the first round, clusters $\{m_1\}$ and $\{m_2\}$ are merged, followed by 
merger of $\{e_3\}$ and $\{m_6\}$ in the second round, and so on. Note that in 
fifth round, clusters $c_1 = \{m_4, e_2\}$ has higher affinity with 
$c_2=\{m_1, m_2, m_3, e_1\}$ than with $c_3=\{m_5\}$, yet $c_1$ and $c_3$ 
are merged instead of $c_1$ and $c_2$ due to the constraint that we cannot merge 
two clusters which \emph{both} contain an entity.  At the end, every mention is 
clustered together with exactly one entity, and there could be entities present 
as singleton clusters such as $\{e_4\}$ and $\{e_5\}$. 
Note that $m_3$ correctly 
links to its gold entity $e_1$ as a result of being clustered with mentions 
$m_1,m_2$ even though it has higher affinity with entity $e_3:\ w(m_3,e_3) > w(m_3,e_1)$.

\subsection{Affinity Models}
\label{sec:affinity_models}

We parameterize $\psi(\cdot,\cdot)$ and $\phi(\cdot,\cdot)$ using two separate deep transformer encoders \cite{vaswani2017attention} for our mention-mention affinity model and mention-entity affinity model --- specifically we use the BERT architecture \cite{devlin2018bert} initialized using the weights from BioBERT \cite{lee2019biobert}. 

\subsubsection{Mention-Mention Model} The mention-mention model is also a cross-encoder, taking as input a pair of mention in context and producing a single scalar affinity for every pair.  The input tokens take the form:
\begin{align*}
    &\texttt{[CLS]} <m_i> \texttt{[SEP]} <m_j> \texttt{[SEP]} \\
    &\text{where } <m_i> \, := c_l \texttt{[START]} m_i \texttt{[END]} c_r
\end{align*}
where $m_i$ is the mention tokens and $c_l$ and $c_r$ are the left and right context of the mention in the text, respectively.
The \texttt{[START]} and \texttt{[END]} tokens are special tokens fine-tuned to signify the start and end of the mention in context, respectively.
We restrict the length of each input sequence to have a maximum of 256 tokens.
A representations of each mention is computed using the average of the encoder's output representations corresponding to the mention's input tokens. The affinity for a mention pair is computed by concatenating their mention representations
and passing it through a linear layer with a sigmoid activation.

\subsubsection{Mention-Entity Model} The mention-entity affinity model is a cross-encoder model \cite[inter alia]{vig2019comparison,wolf2019transfertransfo,humeau2019poly} and takes as input the concatenation of the mention in context with the entity description. 
The input tokens take the form:
\begin{equation*}
    \texttt{[CLS]} c_l \texttt{[START]} m \texttt{[END]} c_r \texttt{[SEP]} e \texttt{[SEP]}
\end{equation*}

\input{zz_fig_alg_box}

\noindent where the mention in context is the same as in the mention-mention model and $e$ is the description of the entity.
We restrict the length of this input sequence to 256 tokens.
After passing the input sequence through BERT, we transform the output representation corresponding to the \texttt{[CLS]} token with a linear layer with one output unit. This value is finally passed through the sigmoid function to output affinity between the mention and the entity.

%% file: zz_fig_alg_box.tex
\begin{algorithm}[t]
\caption{Clustering Inference for Linking}
\begin{algorithmic}[1]
\STATE \textbf{Input:} $(\cM, \cE, \Gamma, \varphi, \psi)$
\STATE \textbf{Output:} $\{(m_i, \hat{e}_i)\}_{i=1}^{|\cM|}$
\STATE $\rhd$ Construct the graph $G$
\STATE $E = \{\}$
\FOR {$m_i \in \cM$}
    \FOR {$m_j \in \cM^{(D_i)}\setminus\{m_i\}$}
        \STATE $E = E \cup \{(m_i, m_j, \varphi(m_i, m_j))\}$
    \ENDFOR
    \FOR {$e_l \in \Gamma(m_i)$}
        \STATE $E = E \cup \{(m_i, e_l, \psi(m_i, e_l))\}$
    \ENDFOR
\ENDFOR
\STATE Construct $G = (V, E)$ from edge set $E$
\STATE Let $S$ be the edges sorted in descending order
\STATE $\rhd$ Cluster nodes of $G$ under linking constraint
\STATE $\hat{\cC} = \{\{v\} | v \in V\}$
\FOR {$(s, t) \in S$} 
    \IF {$\hat{\cC}(s) \cap \cE = \emptyset$ or $\hat{\cC}(t) \cap \cE = \emptyset$}
        \STATE $\hat{\cC} = \hat{\cC} \setminus \{\hat{\cC}(s), \hat{\cC}(t)\}$
        \STATE $\hat{\cC} = \hat{\cC} \cup \{\hat{\cC}(s) \cup \hat{\cC}(t)\}$
    \ENDIF
\ENDFOR
\STATE $\rhd$ Make linking decisions based on clustering
\STATE $L = \{\}$
\FOR{$C \in \hat{\cC}$}
    \STATE $M = C \cap \cM$
    \STATE $\{\hat{e}\} = C \cap \cE$
    \FOR{$m \in M$}
        \STATE $L = L \cup \{(m, \hat{e})\}$
    \ENDFOR
\ENDFOR
\STATE \textbf{return} $L$
\end{algorithmic}
\label{alg:clustering}
\end{algorithm}

%% file: 05_training.tex
\section{Training}
\label{sec:train}

In this section, we explain the training procedure for the affinity models $\varphi(\cdot,\cdot)$ and $\psi(\cdot,\cdot)$  used by the clustering inference procedure. We train the mention-mention and mention-entity models independently in a way that allows the affinities to be comparable when performing inference.

We use triplet max-margin based training objectives to train both models. The most important aspect of our procedure is how we pick negatives during training. For the mention-entity model, we restrict our negatives to be from the candidate set. For the mention-mention model, we restrict our negatives to come from mentions within the same document. From these sets of possible negatives we choose the top-$k$ most offending ones according the instantaneous state of the model -- i.e. the negatives with highest predicted affinities according to the model at that point during training. The following sections detail the training procedures for both models.

\subsection{Mention-Mention Affinity Training}
To train the mention-mention affinity model we use a variant of the maximum spanning tree (MST) supervised single linkage clustering algorithm presented in \citet{yadav2019supervised}.
Let $\cM^{(D)}_{e_l} = \{m \in \cM^{(D)} \,|\, \cE(m) = e_l\}$ be the set of mentions referring to entity $e_l$ in any one document and the set of ground truth clusters be represented by $\cC^* = \{\cM^{(D)}_{e_l} \,|\, e_l \in \cE\}$.
Let $P$ be the set of positive training edges: the edges of the MST of the complete graph on the cluster $C \in \cC^*$. Let 
$N_\varphi(m_*)$ be the $k$-nearest within document negatives to the anchor point $m_{*} \in C$ according to the current state of the model during training.
The objective of this training procedure is to minimize the following triplet max-margin loss\footnote{Define $[x]_+ = \max(x, 0)$} with margin $\mu$ for each cluster $C \in \cC^*$:
\begin{equation*}
    \cL_\varphi(\theta; C) \! = \!\!\! \sum_{m_{*}, m_{+} \in P} \; \sum_{m_{-} \in N_\varphi(m_{*})} \!\!\!\! \ell_{\varphi,\mu}(m_{*}, m_{+}, m_{-}),
\end{equation*}
\begin{equation*}
          \text{where } \ell_{\varphi, \mu}(a, p, n) = [\varphi(a, n) - \varphi(a, p) + \mu ]_{+}.
\end{equation*}

\subsection{Mention-Entity Affinity Training}
For the mention-entity model, we use a triplet max-margin based objective with margin $\mu$ where the anchor is a mention $m$ in the training set, the positive is the ground truth entity $e_{+} = \cE(m)$, and the negatives are chosen from the candidate set $\Gamma(m)$. Denote the $k$ most offending negatives according to the current state of the model during training as $N_\psi(m) \subseteq \Gamma(m) \setminus \{\cE(m)\}$. Formally, the loss is
\begin{equation*}
    \cL_\psi(\theta; \cM)\! = \sum_{m,\, e_+} \; \sum_{e_- \in N_\psi(m)} \ell_{\psi, \mu}(m, e_{+}, e_{-}),
\end{equation*}
\begin{equation*}
          \text{where } \ell_{\psi, \mu}(a, p, n) = [\psi(a, n) - \psi(a, p) + \mu ]_{+}.
\end{equation*}

%% file: 07_experiments.tex
\section{Experiments}
\label{sec:experiments}

\input{zz_fig_examples}

 We evaluate on biomedical entity linking using the MedMentions  \cite{mohan2019medmentions} and BC5CDR \cite{li2016biocreative} datasets.  We compare to state-of-the-art methods. We then analyze the performance of our method in more detail and provide qualitative examples demonstrating our approaches' ability to use mention-mention relationships to improve candidate generation and linking.

\subsection{MedMentions}
\label{sec:dataset}
MedMentions is a publicly available\footnote{\url{https://github.com/chanzuckerberg/MedMentions}} dataset consisting of the titles and abstracts of 4,392 PubMed articles. 
The dataset is hand-labeled by annotators and contains labeled mention spans and entities linked to the 2017AA full version of UMLS. 
Following the suggestion of \citet{mohan2019medmentions}, we use the ST21PV subset, which restricts the entities linked in documents to a set of 21 entity types that were deemed most important for building scientific knowledge-bases. We refer the readers to \citet{mohan2019medmentions} for a complete analysis of the dataset and provide a few important summary statistics here. The train/dev/test split partitions the PubMed articles into three non-overlapping groups. This means that some entities seen at training time will appear in dev/test and other entities will appear in dev/test but not at training time. In fact, a large number of entities that appear in dev/test time are unseen at training, about 42\% of entities. See Table~\ref{tab:dataset_stats} for split details and statistics.

Previous work has evaluated on MedMentions using unfairly optimistic candidate generation settings such as using only 10 candidates including the ground truth \cite{zhu2019latte} or restricting candidates to entities appearing somewhere in the MedMentions corpus \cite{murty2018hierarchical}. We instead work in a much more general setting where all entities in UMLS are considered at candidate generation time and the generated candidates might not include the ground truth entity.

\subsection{BC5CDR}
\label{subsec:bc5cdr}
BC5CDR \cite{li2016biocreative} is another 
entity linking benchmark in the biomedical domain.
The dataset consists of 1,500 PubMed articles annotated
with labeled disease and chemical entities. Unlike MedMentions, which contains 21 types of entities, this dataset contains just two types. These chemical and disease mentions are labeled with entities from MeSH\footnote{https://www.nlm.nih.gov/mesh}, a much smaller biomedical KB than UMLS. See Table~\ref{tab:dataset_stats} for split details and statistics.

\input{zz_fig_dataset_stats}

\subsection{Preprocessing}

The MedMentions ST21PV corpus is processed as follows: (i) Abbreviations defined in the text of each paper are identified using AB3P \cite{Sohn2008AbbreviationDI}. Each definition and abbreviation instance is then replaced with the expanded form. (ii) The text of each paper in the corpus is tokenized and split into sentences using CoreNLP \cite{manning-EtAl:2014:CoreNLP}. (iii) Overlapping mentions are resolved by preferring longer mentions that begin earlier in each sentence, and mentions are truncated at sentence boundaries. This results in 379 mentions to be dropped from the total of 203,282. (iv) Finally, the corpus is saved into the IOB2 tag format. 
The same preprocessing steps are used for BC5CDR, except overlapping mentions are not dropped.

\subsection{Candidate Generation}
 For both datasets, we use a character $n$-gram TF-IDF model to produce candidates for all of the mentions in all splits. The candidate generator utilizes the 200k most frequent character $n$-grams, $n \in \{2 \ldots 5\}$ and the 200k most frequent words in the names in $\cE$ to produce sparse vectors for all of the mentions and entity descriptions (which in our case is the canonical name, the type, and a list of known aliases and synonyms).  Table~\ref{tab:cand_gen} provides candidate generation results for each dataset. The results report the average recall@$K$ at different numbers of candidates ($K$), i.e., whether or not the gold entity is top $K$ candidates for a given mention. 
 

\subsection{Training and Inference Details}
Our model contains 220M parameters, the majority of which are contained within the two separate BERT-based models.  We optimize both the models with mini-batch stochastic gradient descent
using the Adam optimizer \cite{kingma2014adam} with recommended learning rate of 5e-5 \cite{devlin2018bert} with no warm-up steps. We accumulate gradients over all of the triples for a batch size of 16 within document clusters. We compute the top-$k$ most offending negatives on-the-fly for each batch by running the model in inference mode proceeding each training step. 
Training and inference are done on a single machine with 8 NVIDIA 1080 Ti GPUs. We train our model on MedMentions for two epochs and BC5CDR for four epochs. Training takes approximately three days for MedMentions and one day for BC5CDR.
Clustering-based inference takes about three hours for MedMentions and one hour for BC5CDR. Code and data to reproduce experiments will be made available. 

\input{zz_fig_all_results}

\subsection{Results}

We compare our clustering-based inference procedure,  which we refer to our approach as \our,
to a state-of-the-art independent 
inference procedure, \independent, which is the zero-shot architecture proposed by \citeauthor{logeswaran2019zero}. This
same model is used as the 
mention-entity affinity model used
in our approach. We also compare to 
to the state-of-the-art model \taggerOne \cite{leaman2016taggerone} on both MedMentions and BC5CDR.

Table~\ref{tab:all_results} shows performance of 
\taggerOne, \independent, and \our inference procedure on MedMentions and BC5CDR. We report results using the gold mention segmentation (rather than end-to-end) to focus on the performance of each model in terms of linking rather than confounding the performance by including segmentation. Due to \taggerOne's joint entity recognition, typing, and linking architecture, we cannot make predictions for gold mention boundaries without also using their gold types. And so to have a fair comparison to TaggerOne, we provide the gold mention boundaries and types to each system and report these results as well.
 
We use \emph{seen} and \emph{unseen} to refer to the sets of mentions whose ground truth entities are seen and unseen at training, respectively. Note that even if a mention is in the subset of mentions referred to as \emph{seen}, it does not mean that we have seen the particular surface form before in the training set, merely that we have seen other mentions of that particular entity.

 On MedMentions, when the models are provided with only the gold mention span, \our inference procedure outperforms \independent 
 by 1.3 points of accuracy, and we see improvements in accuracy for both seen and unseen entities. When the models are additionally provided with the gold type, we see substantial improvements in accuracy for both \independent and \our over \taggerOne, namely 3.0 and 5.3 points of improvement, respectively.
 
  On BC5CDR, when the models are provided with only the gold mention span, \our inference procedure outperforms \independent by 0.4 points of accuracy, and we see improvements in accuracy for both seen and unseen entities. When the models are additionally provided with the gold type, we see improvements in accuracy for both \independent and \our over \taggerOne, namely 1.6 and 1.9 points of improvement, respectively.

 Observe that the candidate generation results are drastically different for the two datasets (Table~\ref{tab:cand_gen}). We posit that the ability to generate correct candidates correlates with the relative difficultly of the linking task on each dataset, respectively.

\subsection{Analysis: Recovering from Poor Candidate Generation}

We hypothesize that our clustering-based inference procedure
would allow for better performance on mentions for which candidate 
generation is difficult. Observe that while the performance
of the independent model is upper bounded by the recall of candidate generation, this is not an upper bound for the clustering-based model. The clustering-based model can allow mentions that have no suitable candidates to link to other mentions in the same document. We report the accuracy of both systems with respect to whether or not the ground truth entity is in each mentions' list of candidates. 

The accuracy for each system and each partition of mentions is shown in Table~\ref{tab:cand_gen_analysis}. Observe that our approach offers a large number of mentions a correct resolution, when the independent model could not link them correctly due to the ground truth entity being missing from the candidate list. Additionally, it can be seen that \our does sacrifice some performance in comparison to \independent, but more than makes up for it in the case where the ground truth entity is not in the candidate set.

\input{zz_fig_candidate__gen_analysis}
\input{zz_fig_amb_mentions}

\subsection{Analysis: Handling Ambiguous Mentions}

We also hypothesize that for mentions which are highly ambiguous and could refer to many different entities, such as common nouns like \emph{virus}, \emph{disease}, etc, the clustering-based inference should offer improvements. Table~\ref{tab:amb_men} shows that our approach is able to correctly link more ambiguous mentions compared to independent model\footnote{These are: \emph{activation,  activity, a, b, cardiac, cells,  clinical, compounds, cr, development, disease, function, fusion, inhibition, injuries, injury, liver, management, methods, mice, model, pa, production, protein, regulation, report, responses, response, r, screening, stress, studies, study, treatment}}. Figure~\ref{fig:examples} shows two examples from this subset where \our inference is able to make the correct linking decision and \independent is not.


 \input{zz_fig_cand_gen}

%% file: zz_fig_examples.tex
\begin{figure*}
    \centering
    \includegraphics[width=\textwidth]{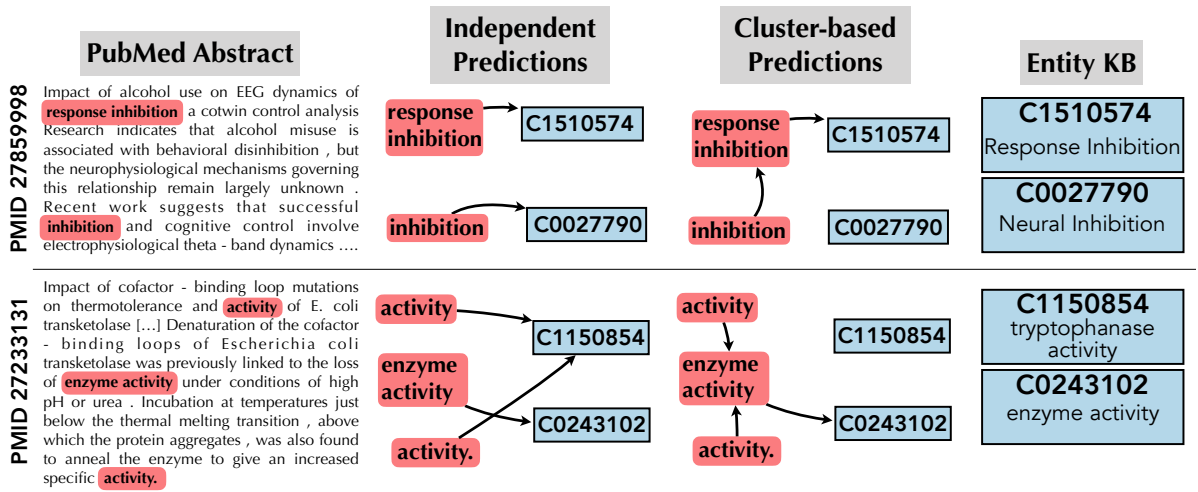}
    \caption{\textbf{Example predictions on Ambiguous Mentions}. Here we show two example outputs for highly ambigous mention surface forms (\texttt{inhibition} and \texttt{activity}). The independent model incorrectly makes predictions on these surface forms. The clustering-based model is able to have each ambiguous mention link to a less ambiguous mention in the same abstract and thereby make correct predictions.}
    \label{fig:examples}
\end{figure*}

%% file: zz_fig_dataset_stats.tex


\begin{table}[]
    \footnotesize
    \centering
    \begin{tabular}{l  c c c c c c  }
    \toprule
    & \multicolumn{3}{c}{\bf MedMentions} & \multicolumn{3}{c}{\bf BC5CDR} \\
                            & \bf Train     & \bf Dev       & \bf Test & \bf Train     & \bf Dev       & \bf Test \\
    \midrule
    \bf $|\mathcal{M}|$  & 120K   & 40K    & 40K    & 18K   & 934    & 10K\\
    \bf $|\mathcal{E}|$  &   19K & 9K  & 8K    &   2K & 281 & 1K  \\
    \bf $\%$ seen   &  100        & 57.7 & 57.5  &  100        & 80.1 & 64.8 \\
    \bottomrule
    \end{tabular}
    \caption{\textbf{Linking Datasets}. Statistics of each dataset, including the percent of ground truth entities seen during training ($\%$ seen).}
    \label{tab:dataset_stats}
\end{table}

%% file: zz_fig_all_results.tex
\begin{table*}[]
    \centering
    \begin{tabular}{lccc|ccc}
    \toprule
    &  \multicolumn{3}{c|}{\bf MedMentions} &  \multicolumn{3}{c}{\bf BC5CDR} \\
    & \bf Overall & \multicolumn{2}{c|}{\bf Acc. on } &  \bf Overall & \multicolumn{2}{c}{\bf Acc. on } \\
    & \bf Acc.   & \bf Seen & \bf Unseen & \bf Acc.   & \bf Seen & \bf Unseen \\
    \midrule
    
    \hspace{0.5cm}\textsc{N-gram TF-IDF} & 50.9  & 50.9 & 51.0 & 86.9 & 89.2 & 74.6 \\
    \hspace{0.5cm}\textsc{BioSyn}\tiny{~\cite{sung2020biomedical}}   & $72.5^\dagger$  & $76.5^\dagger$ & $58.7^\dagger$ & $87.8^\dagger$ & $89.0^\dagger$ & $81.1^\dagger$ \\
    \hspace{0.5cm}\textsc{SapBERT}\tiny{~\cite{liu2020selfalignment}}   & $69.8^\dagger$  & $72.9^\dagger$ & $58.9^\dagger$ & $85.2^\dagger$ & $85.8^\dagger$ & $\textbf{82.0}^\dagger$ \\
    \hspace{0.5cm}\independent\tiny{~\cite{logeswaran2019zero}}   & 72.8  & 75.9 & 61.9 &  90.5  & 94.0 & 73.6  \\
       \hspace{0.5cm}\our (ours)                                     & \textbf{74.1}  & \textbf{77.3} & \textbf{62.9} & \textbf{91.3}  & \textbf{94.9} & 73.8 \\
    \midrule
        w/ Gold Types &  &  &   &  &  &  \\
    \hspace{0.5cm}\textsc{N-gram TF-IDF} & 67.9  & 69.0 & 64.0 & 87.8 & 90.2 & 76.1 \\
    \hspace{0.5cm}\taggerOne\tiny{~\cite{leaman2016taggerone}}    & 73.8  & 78.2 & 58.8 & 89.8 & 91.8 & 79.9 \\
    \hspace{0.5cm}\textsc{BioSyn}\tiny{~\cite{sung2020biomedical}}   & $77.0^\dagger$  & $80.7^\dagger$ & $64.1^\dagger$ & $87.9^\dagger$ & $89.1^\dagger$ & $81.3^\dagger$ \\
    \hspace{0.5cm}\textsc{SapBERT}\tiny{~\cite{liu2020selfalignment}}   & $74.1^\dagger$  & $77.0^\dagger$ & $63.8^\dagger$ & $86.0^\dagger$ & $86.8^\dagger$ & $\textbf{82.0}^\dagger$ \\
    \hspace{0.5cm}\independent\tiny{~\cite{logeswaran2019zero}}   & 76.8  & 79.2 & 68.4 & 90.6 & 94.1 & 73.6 \\
    \hspace{0.5cm}\our (ours)                                     & \textbf{79.1} & \textbf{81.5} & \textbf{70.5} & \textbf{91.4} & \textbf{94.9} & 74.0   \\
    \bottomrule
    \end{tabular}
    \caption{\textbf{Entity Linking Results}. We report linking accuracy on MedMentions and BC5CDR datasets with gold mentions spans and gold mention spans and types. We observe that \our inference provides improved accuracy in each setting with additional improvements seen when gold entity types are provided. ($^\dagger$Hits at one synonym --- multiple entities could be predicted)}
    \label{tab:all_results}
\end{table*}

%% file: zz_fig_candidate__gen_analysis.tex
\begin{table*}[]
    \centering
    \begin{tabular}{l  c  c | c  c }
    \toprule
            & \multicolumn{2}{c}{\bf MedMentions} & \multicolumn{2}{c}{\bf BC5CDR} \\
            & $\ \Ecal(m) \in \Gamma(m)$ & $\ \Ecal(m) \not \in \Gamma(m)$ & $\Ecal(m)\in \Gamma(m)$ & $\ \Ecal(m) \not \in \Gamma(m)$  \\
    \midrule
   
    \textsc{\quad Independent} & \bf 85.3 & 0.0 & \bf 95.5 & 0.0 \\
    \quad \our  &  84.5 & \bf 13.9 & 95.3 & \bf 14.9 \\
    \midrule
    w/ Gold Types & & & & \\
    \textsc{\quad Independent} & \multirow{ 1}{*}{\bf 90.0} & \multirow{ 1}{*}{0.0} & \multirow{ 1}{*}{\bf 95.7} & \multirow{ 1}{*}{0.0} \\
   \quad \our  &  \multirow{ 1}{*}{89.3} & \multirow{ 1}{*}{\bf 19.3} & \multirow{ 1}{*}{95.4} & \multirow{ 1}{*}{\bf 15.9} \\
    \bottomrule
    \end{tabular}
    \caption{\textbf{Performance when Candidate Generation Fails}. We report the accuracy of each method on mentions for which the ground truth entity is in the candidate list ($\Ecal(m) \in \Gamma(m)$) and is not in the list ($\Ecal(m) \not \in \Gamma(m)$). We observe that our proposed approach is able to perform reasonably well even when candidate generation fails.}
    \label{tab:cand_gen_analysis}
\end{table*}

%% file: zz_fig_amb_mentions.tex
\begin{table}[]

    \centering
    \begin{tabular}{l c }
    \toprule
            & \bf Accuracy \\
    \midrule
    \textsc{Independent} \tiny{~\cite{logeswaran2019zero}} & 71.91  \\
    \our (ours)  & \bf 73.03  \\
    \bottomrule
    \end{tabular}
    \caption{\textbf{Performance on Ambiguous Mentions} We select mentions for which the surface form is labeled 10 or more different entities in MedMentions and measure performance on instances of these surface forms on the test data. We observe that \our is able to more accurately link these mentions. Figure \ref{fig:examples} shows examples of these mentions. }
    \label{tab:amb_men}
\end{table}

%% file: zz_fig_cand_gen.tex
\begin{table}[]
    \centering
    \begin{tabular}{c  c c  }
    \toprule
        \bf Recall@     & \bf BC5CDR   & \bf MedMentions \\
    \midrule
    1 & 86.9 & 50.8 \\
    2 & 89.4 & 63.8 \\
    4 & 91.1 & 73.4 \\
    8 & 92.1 & 79.2 \\
    16 & 93.1 & 82.3 \\
    32 & 94.3 & 84.6 \\
    64 & 94.9 & 85.3 \\
    \bottomrule
    \end{tabular}
    \caption{\textbf{Candidate Generation Recall}. Recall is measured by whether or not the ground truth entity is in the top K candidate entities for the given mention. We report the micro average recall over all mentions.}
    \label{tab:cand_gen}
\end{table}


%% file: 08_related_work.tex
\section{Related Work}

Entity linking is widely studied and often focused on linking
mentions to Wikipedia entities (also known as Wikification) \cite{mihalcea2007wikify,cucerzan2007large,milne2008learning,hoffart2011robust,ratinov2011local,cheng2013relational}. 
Entity linking is often done independently for each mention in the document \cite{ratinov2011local,raiman2018deeptype} or by modeling dependencies between predictions of entities in a document \cite{cheng2013relational,ganea2017deep,le2018improving}.

In the biomedical domain, Unified Medical Language System (UMLS) is often used as a knowledge-base for entities \cite{mohan2019medmentions,leaman2016taggerone}. While UMLS is a rich ontology of concepts and 
relationships between them, this domain is low resource compared to Wikipedia with respect
to number of labeled training data for each entity mention. This leads to a zero-shot setting
in datasets such as MedMentions \cite{mohan2019medmentions} where new entities are seen at test time. Previous work has addressed this zero-shot setting using models of the type hierarchy \cite{murty2018hierarchical,zhu2019latte}. This previous work \cite{murty2018hierarchical,zhu2019latte} uses an unrealistic candidate generation setting where the true positive candidate is within the candidate set and/or entities are limited to those in the dataset rather than those in the knowledge-base.

Mention-mention relationships are also explored in \cite{le2018improving} which extends the pairwise CRF model \cite{ganea2017deep} to use mention-level relationships in addition to entity relationships. These works use attention in a way to build the context representation of the mentions. However, as mentioned by \citet{logeswaran2019zero} is not well suited for zero-shot linking.

Coreference (both within and across documents) has also been explored by past work \cite{dutta2015c3el}. This work 
uses an iterative procedure that performs hard clustering for the sake of aggregating the contexts of entity mentions. 
\citet{durrett-klein-2014-joint} presents a CRF-based model for joint NER, within-document coreference, and linking. They show that jointly modeling these three tasks improves performance over the independent baselines. This differs from our work since we do not require coreference decisions  to be correct in order to make correct linking decisions. 
Other work performs joint entity and event coreference \cite{barhom-etal-2019-revisiting} without  linking.

%% file: 09_conclusion.tex
\section{Conclusion}
In this work, we presented a novel clustering-based inference procedure which enables joint entity linking predictions. We evaluate the effectiveness of our approach on the two biomedical entity linking datasets, including the largest publicly available dataset. We show through analysis that our approach is better suited to link mentions with ambiguous surface forms and link mentions where the ground truth entity is not in the candidate set.

%% file: 11_ethics.tex
\section{Ethical Considerations}
Entity linking is a task with the intention of providing useful information when building a semantic index of documents. This semantic index is a core component of systems which allow users to search, retrieve, and analyze text documents. In our specific case, we are interested in building semantic indexes of scientific documents where the end user would be scientists and researchers. The goal is to help them navigate the vast amount of literature and accelerate science. This being said, users need to take the outputs of such a system as suggestions and with the potential that the information is incorrect. Researchers must be aware that the system is not perfect and they should not jump to any conclusions especially about important decisions. Additionally, the researcher can always verify the decisions being made by the system.

While this paper focuses on biomedical entity linking, this technique could be extended to other domains. In such other domains, users might not have as much expertise, but the user is still responsible for making decisions on their own, since the system is not perfect. In addition, the system developers and designers need to be aware of their particular application to ensure to mitigate harm which could come from such a system. For example, in any application that deals with personalized data, we need to be wary of the potential outcomes which could come from an entity linking based system or semantic index, such as privacy or other potential malicious behaviour or unforeseen consequences due to the decisions being made by the system.

%% file: 10_ack.tex
\section*{Acknowledgements}

We thank members of UMass IESL and NLP groups for helpful discussion and feedback. This work is funded in part by the Center for Data Science and the Center for Intelligent Information
Retrieval, and in part by the National Science Foundation under Grants No. 1763618, and in part by the Chan Zuckerberg Initiative under the project Scientific Knowledge
Base Construction. 
The work reported here was performed in part by the Center for Data Science and the Center for Intelligent Information Retrieval, and in part using high performance computing equipment obtained under a grant from the Collaborative R\&D Fund managed by the Massachusetts Technology Collaborative.
Rico Angell is supported by the National Science Foundation Graduate Research Fellowship under Grant No. 1938059.
Any opinions, findings and conclusions or recommendations expressed in this
material are those of the authors and do not necessarily reflect those of the sponsor.